\documentclass{article}

\usepackage{arxiv}

\usepackage[utf8]{inputenc} % allow utf-8 input
\usepackage[T1]{fontenc}    % use 8-bit T1 fonts
\usepackage{hyperref}       % hyperlinks
\usepackage{url}            % simple URL typesetting
\usepackage{booktabs}       % professional-quality tables
\usepackage{amsfonts}       % blackboard math symbols
\usepackage{amsmath}
\usepackage{nicefrac}       % compact symbols for 1/2, etc.
\usepackage{microtype}      % microtypography
\usepackage{lipsum}
\usepackage{graphicx}
\usepackage{comment}
\usepackage[noend]{algpseudocode}
\usepackage{algorithmicx}
\usepackage{algorithm}
\graphicspath{ {./} }
\title{The Brownian motion in the transformer model}

\author{
  Yingshi Chen\thanks{https://github.com/closest-git/} \\
  \texttt{gsp@grusoft.com} \\
}

\begin{document}
\maketitle 
\begin{abstract}
 Transformer is the state of the art model for many language and visual tasks. In this paper, we give a deep analysis of its multi-head self-attention (MHSA) module and find that: 1) Each token is a random variable in high dimensional feature space. 2) After layer normalization, these variables are mapped to points on the hyper-sphere. 3) The update of these tokens is a Brownian motion. The Brownian motion has special properties, its second order item should not be ignored. So we present a new second-order optimizer(an iterative K-FAC algorithm) for the MHSA module.   
  
  In some short words: All tokens are mapped to high dimension hyper-sphere. The Scaled Dot-Product Attention $softmax(\frac{\mathbf{Q}\mathbf{K}^T}{\sqrt{d}})$ is just the Markov transition matrix for the random walking on the sphere. And the deep learning process would learn proper kernel function to get proper positions of these tokens. The training process in the MHSA module corresponds to a Brownian motion worthy of further study.
\end{abstract}

% keywords can be removed

\section{Introduction}
Transformer model is one of the greatest innovation since convolution network, which brings surprising best results in many NLP and vision tasks. It's the key component of BERT\cite{devlin2018bert}, ViT\cite{dosovitskiy2020vit} and many other state of the art(SOTA) models. But most study is on the improvement its performance. The essence of this model needs to be analyzed, especially the secret of self-attention mechanism. Only in this way can we continue to improve this model and make further development. In this paper, we use random walking and Brownian motion to explain the multi-head self-attention (MHSA) module, which is the key component of transformer. We reveal its intrinsic connection with Brownian motion\cite{wiki_brownian}, which is a powerful tool to analysis various natural and social phenomena.

\subsection{Great success of Transformer/BERT with no clear explanation} 
BERT(Bidirectional Encoder Representations from Transformers) \cite{devlin2018bert} gives state-of-the-art performance on many NLP tasks. Inspired by its great success, more transformer based model are proposed: Transformer-XL\cite{dai2019transformer}, GPT-2\cite{radford2019better}, XLNet\cite{yang2019xlnet}, RoBERTa\cite{liu2019roberta}, ALBERT\cite{lan2019albert}, DistilBERT\cite{sanh2019distilbert}...  These models propose some new structures to improve its performance or reduce the number of parameters.

Why BERT is so successful and how to explain? There have been some attempts to answer, but they are mainly based on experiments and observations and lack of in-depth and clear explanation. In this survey of MIT \cite{rogers2020primer}, they listed over 150 studies of BERT model and get the following conclusion - "BERTology has clearly come a long way, but it is fair to say we still have more questions than answers about how BERT works."  \cite{kovaleva2019revealing} provides valuable insights into what happens during fine-tuning, but the scope of their conclusions is limited: no clear linguistic phenomena is being captured by attention. 

Especially, attention mechanism(key module of transformer) is getting more and more research, and causes a lot of controversy. For example, there are a series of articles to discuss the "Attention / Explanation" problem. At first, \cite{jain2019attention} declares that "Attention is not explanation", attention weights do not provide meaningful “explanations" for predictions. \cite{wiegreffe2019attention} declares that "Attention is not not explanation".They propose four alternative tests which would get meaningful interpretation from attention mechanisms.And then \cite{grimsley2020attention} gives the opposite conclusion. They asserts the impossibility of causal explanations from attention layers over text data. Instead of answering yes or no, \cite{bastings2020elephant} put forward a new plan. Their views are very interesting: ”While attention conveniently gives us one weight per input token and is easily extracted, it is often unclear toward what goal it is used as explanation ... input saliency methods are better suited, and that there are no compelling reasons to use attention".

These seemingly contradictory conclusions reflect the lack of clear understanding now, and it is necessary to reveal the deep secret of transformers, especially the secret of attention mechanism. And based on our theoretical analysis, the attention mechanism(MHSA) in the transformer is really reliable and meaningful.

\subsection{Vision Transformer(ViT) }
Inspired by the great success of BERT, ViT \cite{dosovitskiy2020vit,wu2020visual,tolstikhin2021mixer,steiner2021augreg} also use transformers. A simple and efficient technique in \cite{dosovitskiy2020vit} is to take a small pixel as a token, and then directly train it with the existing BERT model. They interpret an image as a sequence of patches and process it by a standard Transformer encoder as used in NLP. The high accuracy reported in \cite{dosovitskiy2020vit}
reflects the great power of transformer model. Even this simple technique would beat many deep CNN models, which have been studied and optimized by many years and groups. 

To explain the success of these ViT model, we still need to understand the nature of the Transformers. This is what this paper focuses on. 

\begin{comment}
However, details of the Transformer architecture –such as the use of non-overlapping
patches– lead one to wonder whether these networks are as robust\cite{bhojanapalli2021understanding}. 
\end{comment}

\subsection{Our contributions}
We present some novel understandings of transformers, especially the Brownian motion in the MHSA module. In some short words: All tokens are mapped to high dimension hyper-sphere. The Scaled Dot-Product Attention $softmax(\frac{\mathbf{Q}\mathbf{K}^T}{\sqrt{d}})$ is just the Markov transition matrix for the random walking on the sphere. And the deep learning process would learn proper kernel function to get proper positions of these tokens. Our analysis points out a new direction. That is, analysis and improve the transformer model with mathematical model, which has a solid theoretical foundation and strong description ability.
\begin{comment}
That is, some layers are the interaction between voxels, and others are global operation applied to all voxels. From the operator splitting method, these two are total different operators and we could solver it by three-stage strang-scheme. So we design a compact model with three blocks, The middle one is MHSA block, which reflects the interaction between different voxels.And the other two is high-width MLP structure, which represents the global transformation. The experiment on the cifar data-sets shows its advantages.
\end{comment}

\section{Background and notation}
\label{sec:Background}
In this section, we give some background knowledge which are needed in the following analysis and derivation. 

\subsection{Kernel function}    \label{back:kernel}
Kernels function has been successfully applied to many methods, especially support-vector machine(SVM) \cite{scholkopf2002learning}. In these scenarios, kernel is a simple and effective description of distance or similarity between two objects. No limits on the format of objects and the space they exist in. For two vectors or tensors, the linear kernel(or dot-product) is widely used. That is, the dot-product is just a measure of similarity of tow vectors.This kernel could be implemented efficiently based on matrix multiplication. For example, for a batch of tensor in deep learning, we would use high-performance tensor production to get all similarity between all vectors.

The Gaussian kernel is a non-linear function of tow vectors' euclidean distance.
\begin{equation} \label{F:kernel_gaussian}
K(x,y)=\mathrm{exp}\left ( -\frac{\left \| x-y \right \|^2}{2\sigma^2} \right )
\end{equation}

The connectivity between two data points, x and y, is
defined as the probability of jumping from x to y in one
step of the random walk. And this probability is the normalize kernel function. For example, the softmax normalized kernel function.

\subsection{Markov chain and Transition matrix } \label{back:markov}
A discrete-time Markov chain is a sequence of random variables $ X_1, X_2, X_3, \cdots $ with the Markov property, namely that the probability of moving to the next state depends only on the present state and not on the previous states:
\begin{equation}
Pr( X_{n+1}=x\mid X_{1}=x_{1},X_{2}=x_{2},\ldots ,X_{n}=x_{n})=Pr(X_{n+1}=x\mid X_{n}=x_{n} )
\end{equation}

A transition matrix $M$ is a square matrix used to describe the transitions of a Markov chain . 

\begin{equation}
M=
\begin{bmatrix} 
	M_{1,1} & \cdots & M_{1,j} & \cdots & M_{1,n} \\
	& & \vdots   \\
	M_{i,1} & \cdots & M_{i,j} & \cdots & M_{i,n} \\
	& & \vdots   \\
	M_{n,1} & \cdots & M_{n,j} & \cdots & M_{n,n} \\
\end{bmatrix}
\end{equation}

Each element $ M_{i,j} $ is a non-negative real number representing a probability $ Pr(j|i) $ of moving from state $i$ to state $j$.
Each row summing of $M$ is 1: $ \sum_{j} M_{i,j}=1 $.

The probability transition of from any state to another state in  k steps is given by $M^k$.

\subsection{Random walking in graph and diffusion equation}

A classical Markov chain is Random walking in graph. Let $ G = (V, E)$ be a connected graph with $|V|$ nodes and $|E|$ edges. Let's start from a random node $v_0$ with some initial distribution $P_0$; after $i$ steps we are at
a node $v_i$, then walk to neighbor $v_j$ with probability $p_{ij}$. This process
described by $ {v_i: i = 0, 1, . . .} $ is a Markov chain\cite{masuda2017random}
\cite{lovasz1993random}. 

Let $P_t: Prob(v_t = i)$ is the distribution of $v_t$, and $ = Pr(j|i) \: \forall i,j\in V$ is the transition matrix, then
\begin{equation}
P_{t+1}=M^TP_t
\end{equation}

The continuum limit of the random walk model is known as "diffusion". And the diffusion equation \cite{ben2000diffusion} is just
\begin{equation}
\frac{\partial P}{\partial t} = D\frac{\partial^2P}{\partial x^2}
\end{equation}
where $D$ is the diffusion coefficien

The distribution $P_t$ is very interesting and worthy of further study. It would show the geometric structures of $X$ at various scales \cite{coifman2006diffusiond}.  \cite{coifman2006diffusiond} proposed a diffusion map framework to computes a family of embedding of a data set into Euclidean space. In this diffusion framework: running the chain forward in time (taking larger and larger powers of {M}) reveals the geometric structure of {X} at larger and larger scales . Just like the rendering in Chinese ink painting.

\begin{comment}
\subsection{reaction operator}
The key difference between diffusion map and the transformer is the reaction operator. The diffusion map mehtod would embedd to k larget eigenvector, and transformer would embedd to a fix-length vector. So the transformer method has much higher accuracy than the eigenvector model.
\end{comment}

\subsection{Brownian motion and  Itō's Calculus}
Mathematically Brownian motion $\boldsymbol{B}_t $ is a set of random variables, one for each value of the real variable $t$ in the interval $[0, T]$. This collection has the following properties:
\begin{itemize}
\item $\boldsymbol{B}_t$ is continuous in the parameter $t$, with $B_0 = 0$.
\item For each $t$, $B_t$ is normally distributed with expected value 0 and variance t, and they are independent of each other.
\item For each $t$ and $s$ the random variables $B_{t+s} - B_s$ and $B_s$ are independent. And $B_{t+s} - B_s$ has variance $t$.
\end{itemize}

 For a function $f(t, B_t)$ depends both on some Brownian motion $\boldsymbol{B}_t $ and real variable $t$, the Taylor expansion of $df$ is 
 \begin{equation} \label{F:df_taylor}
df=\frac{\partial f}{\partial t} dt + \frac{\partial f}{\partial B_t} dB_t + \frac{1}{2}\frac{\partial^2 f}{\partial t^2}\left ( dt \right )^2 + \frac{\partial^2 f}{\partial t \partial B_t} dtdB_t + \frac{1}{2}\frac{\partial^2 f}{\partial B_t^2} \left (dB_t  \right )^2 + higher\,  oder\,  terms
\end{equation}
The key difference between $\boldsymbol{B}_t $ and variable $t$ is that the second order item $dB_t^2$ cannot be ignored, or more precisely $dB_t^2 = dt$\cite{wilmott1995mathematics}. So

\begin{equation} \label{F:ito}
\: Ito’s\:lemma: \: df=\left ( \frac{\partial f}{\partial t}+\frac{1}{2}\frac{\partial^2 f}{\partial B_t^2} \right )dt+ \frac{\partial f}{\partial B_t} dB_t 
\end{equation}

The formula \ref{F:ito} is just the famous Ito’s lemma \cite{wilmott1995mathematics}.

For a simplified function $f(B_t)$ which only depends on a Brownian motion $\boldsymbol{B}_t $, we get 
\begin{equation} \label{F:ito_simple}
 df = \frac{1}{2}\frac{\partial^2 f}{\partial B_t^2} dt + \frac{\partial f}{\partial B_t} dB_t 
\end{equation}

\subsection{Vector and Matrix Calculus}
\subsubsection{Softmax function and its derivative}
For a vector $[a_1,a_2,\cdots ]^T$ and its softmax function $[S_1,S_2,\cdots ]^T$, the per-element formula is:
\begin{equation} \label{F:softmax}
     S_j=\frac{e^{a_j}}{\sum e^{a_k}}
\end{equation}    
\\The derivative of the above formula is:
\begin{equation} \label{F:softmax_grad}
\frac{\partial S_i}{\partial a_j}=\frac{\partial \frac{e^{a_i}}{\sum e^{a_k}}}{\partial a_j}=S_i\left ( \delta _{ij}-S_j \right )
\end{equation}   
where $\delta _{ij}$ the Kronecker delta function, it's $1$ when $i=j$, otherwise is 0. For the detailed proof, please see \cite{EliSoftmax}.

\section{The Brownian motion in the MHSA module}    \label{sec:Brownian motion}

Given training data $D=\left \{ X_i,y_i,i=1,2,\cdots  \right \}$, deep learning method tries to reduce the loss $\mathfrak{L}$ between prediction $\hat{y}=f\left ( \theta : X \right )$ and the target $y$. Each sample $ X_i\in \mathbb{R}^{n\times d} $ contains $n$ tokens $(v_1,v_2,\cdots,v_n)$. Each token is embedded into $d$ dimensional feature space $(v_i^1,v_i^2,\cdots,v_i^d) \in \mathbb{R}^d$. In this section, we would analyze how attention module would update $v_i$ to show its actually a Brownian motion.

\begin{comment}
At each step of optimization process, the parameters $\theta $ would be updated along a search direction $p$ with a small step length $\eta$ 
\cite{nocedal2006numerical,absil2009optimization}:
    \begin{equation}
    \begin{split}
    {\theta}'   &= \theta+\eta p       
    \end{split}
    \end{equation}
\end{comment}

\subsection{Layer normalization and hyper-sphere mapping} \label{sec:normal}
Layer normalization(LN) is an important technique to normalize the distribution of data. For any vector $\boldsymbol{v}$,
\begin{equation} 
\textrm{LayerNorm} \left ( \boldsymbol{v} \right )=\gamma \frac{\boldsymbol{v}-u}{\sigma }+\beta  
\end{equation}
where $u,\sigma$ are the mean and standard deviation, $\gamma,\beta$ are learnable parameters. 

For tensors with multiple dimensions in deep learning, there are do some different implementation in different models. In the practical implementation of transformer/BERT, the normalization always acts on the last dimension of the tensor \cite{xiong2020layer,codeTransformer}. The last dimension corresponds to the token. So the effect of LN is to normalize each token so that it's mean is zero and standard deviation is one. That is, each token is mapped to high dimension sphere with its center in the origin. It's easy to prove that the radius of this sphere is equal to the square root of dimension size \cite{sun2020new}. That is, $\left \| \textrm{LN}(\boldsymbol{v}) \right \|_2=\sqrt{d}$. And more, for any $v_i,v_j$ located in this sphere, its dot product is:
\begin{equation} \label{F:dot_ij}
dot(v_i,v_j)=(2d-\left \| v_i-v_j \right \|^2)/2
\end{equation}

The originally designed Transformer places the layer normalization after MHSA module, which is usually referred to as Post-Layer Normalization (Post-LN Transformer). On the other hand, \cite{xiong2020layer} shows that if the layer normalization is put before MHSA (Pre-LN Transformer), the gradients are well-behaved at initialization. Their experiments show that Pre-LN Transformer make training easier and faster. Pre-LN Transformer would simplify our analysis. That is, all the inputs of MHSA are points on the sphere.

Even in the case of Post-LN Transformer model, most inputs of transformers are still normalized. Because nearly all practical models  stack multiple transformer continuously. Only the the input of first transformer is not normalized. All the input of other transformers 
are still normalized. So the position of LN layer is not a big problem.

So we could assume that all the inputs of MHSA are points on the sphere and formula \ref{F:dot_ij}  always holds.

\subsection{The Brownian motion in MHSA module} \label{sec:brown}
Our analysis is based on the following observations:
\begin{itemize}
\item Each token $v_i$ is a random variable.

As pointed in \cite{vaswani2017attention}, each components of token $v_i$ is independent random variable, so $v_i$ is a random variables in $d$ dimensional feature space. 
\item After layer normalization, these variables are points on the hyper-sphere.

As pointed in section \ref{sec:normal}: $\left \| \textrm{LN}(v_i) \right \|_2=\sqrt{d}$.

\item The update of $v_i$ is a stochastic process. We would  analyze some characteristics of this process in this section.
\end{itemize}
 
Let's start from the original formula in the pioneering paper "Attention is all you need" \cite{vaswani2017attention}. 
\begin{equation} \label{F:normal_QKV}
     Attention(\mathbf{Q},\mathbf{K},\mathbf{V})=softmax(\frac{\mathbf{Q}\mathbf{K}^T}{\sqrt{d}})\mathbf{V}
\end{equation}
where 
\begin{itemize}
\item $V$ is a three dimensional tensor, the first dimension corresponds to the training batch, which usually includes 8,16, ... samples. And each sample $ X_i\in \mathbb{R}^{n\times d} $.
\item $ \mathbf Q = \mathbf K = \mathbf V $ (self attention).
\item d is the dimension of each key. \cite{vaswani2017attention} uses $ \sqrt{d} $ as the scaling factor to reduce the large gradient value to improve the softmax accuracy. And $ \sqrt{d} $ is also the radius of  of the hyper-sphere as pointed in \ref{sec:normal}.
\end{itemize}

This formula is much simpler than various complex models appeared in practical applications, but it does not affect the validity of our analysis, as pointed in later subsection.

Let $\mathbf{P} = softmax(\frac{\mathbf{Q}\mathbf{K}^T}{\sqrt{d_k}})$, then sum of each row in $\mathbf{P}$ is 1. The value in each row corresponds to a probability distribution. So $\mathbf{P}$ is a transition matrix of some Markov Process as shown in \ref{back:markov}. It reflects a random walking in high dimensional space. $P_{i,j}$ is the probability that $v_i$ would walk to $v_j$. As the time and space increments to zero, the limitation of random walking process is just Brownian motion. 
Brownian motion has many special properties, one of which the deep learning algorithm must pay attention to is the treatment of second order term. As point by Ito’s lemma, the second order item should not be ignored. So we propose a new second order optimizer method in the next section.

Let's check the sofmax process to calculate $P_{i,j}$, we would first calculate $p_{i,j} = \mathrm{exp}(\frac{dot(v_i,v_j)}{\sqrt{d}})$,
then $P_{i,j}=\frac{p_{i,j}}{\sum_j p_{i,j}}$.
The $dot$ function in $p_{i,j}$ could be further simplified from formula \ref{F:dot_ij}:

\begin{equation}
\begin{split} \label{F:P_gaussian}
 p_{i,j} &= \mathrm{exp}(\frac{dot(v_i,v_j)}{\sqrt{d}})=\mathrm{exp}(\frac{2d-\left \| v_i-v_j \right \|^2}{2\sqrt{d}})=s\times  \mathrm{exp}(\frac{-\left \| v_i-v_j \right \|^2}{2\sqrt{d}})   \\
     s &= \mathrm{exp}(\sqrt{d})    \\
\end{split}
\end{equation}
$s$ is a constant and would be canceled in $P_{i,j}$. So dropping this constant does not affect the calculation. So
\begin{equation}
p_{i,j} = \mathrm{exp}(\frac{-\left \| v_i-v_j \right \|^2}{2\sqrt{d}}) 
\end{equation}
Compared to the definition of Gaussian kernel in subsection \ref{back:kernel}, we would see that $p_{i,j}$ is a special form of Gaussian kernel.
Based on the above analysis, the MHSA module is a Brownian motion which defined on a special Gaussian kernel function.

\subsection{Sencond order K-FAC optimizer algorithm for the training of MHSA}
As mentioned above, the update of tokens' embedding tensors(feature tensors) in MHSA module is actually Brownian motion. The second order item of Brownian motion should not be ignored. So we should use proper optimization methods which consider this characteristic. The commonly used first-
order methods only use gradient information to update parameters(weights). The second-order term is ignored, which leads to slow convergence process. 

There are some second-order method would use second order item. Most reason is try to get
fast convergence on the addition curvature information. Based on our analysis, the second-order method is not only for fast convergence, but is also for the high accuracy.

The most promising second-order method is Kronecker-factored Approximate Curvature (K-FAC) method
\cite{amari1998natural,amari2000adaptive,martens2020new}. In some large-scale learning problems, K-FAC method needs less training
time than SGD(or other first-order method) to get same accuracy. As the following formula shows, K-FAC method
tries to find the
steepest decent direction direction  in the distribution space, which is guided by additional constraint from KL divergence\cite{kullback1997information}:
\begin{equation}
p=\arg\min_{\textrm{KL}\leq\epsilon}\mathit{\mathcal{L}}\left(\theta+p\right)\label{eq:KL}
\end{equation}
where $p$ is the steepest decent direction, $\theta$ is model's parameters and $ \mathcal{L} $ is the loss
between prediction $\hat{y}=f(\theta:x)$ and the target $y$. $\textrm{KL}\leq\epsilon$ would reduce the variance of distribution between steps.

With additional constraint $\textrm{KL}\leq\epsilon, $ K-FAC method is not only looking for suitable parameters, but also
for the distributions that reflect the essence of the problem more
deeply than parameters.The number and value of the parameters will
vary greatly, but the distribution should be always the same. As the loss gets smaller, the distribution changes smaller and smaller. Let the hessian of KL metric is $G$. Then the second-order search
direction of formula \ref{eq:KL} is:
\begin{equation}
\widetilde{\nabla}_{\theta}\mathcal{L}=G^{-1}\nabla_{\theta}\mathcal{L}
\end{equation}
\cite{watanabe2009algebraic}
pointed that the Fisher information matrix(FIM) is equal to the hessian
matrix of the Kullback--Leibler distance. So we would update the
parameters \ensuremath{\theta} by $F$(Fisher information matrix):

\begin{equation}
\theta'=\theta+\eta F^{-1}\nabla_{\theta}\mathcal{L}
\end{equation}

In practical case of deep learning, the dimension of F is very large.
For example, AlexNet has 60 million parameters and BERT\_large has
340 million parameters. The standard method to get $F^{-1}$ would fail
for such huge matrices or would be very slow. K-FAC method approximates
$F$ as a block-diagonal matrix where each block is an inverse of tiny Kronecker factors, then get the inverse of $F$ very quickly. 

The following is an iterative K-FAC algorithm. It uses conjugate
gradient method\cite{saad2003iterative,van1983matrix} to update the tokens’ embedding tensors in MHSA module. This CG-FAC method
is matrix-free, that is, no need to generate the FIM matrix, also
no need to generate the Kronecker factors. For the detail, please see \cite{chen2021iterative}. 
\begin{algorithm}[H]
\caption{CG-FAC method to update token's embedding tensor with at most m iterations}
\hspace*{0.02in} \textbf{Input:} 

$\qquad n$: The parameters size in $i^{th}$ layer

$\qquad a_{i-1}$: The activation of the $(i-1)^{th}$ layer

$\qquad g_{i}$: The gradient of output in the $i^{th}$ layer

$\qquad\mathbf{b}=\nabla L$: The gradient of parameters in $i^{th}$
layer

$\qquad F_{\gamma}=\hat{F_{i}}+\gamma I$: The fisher information
matrix (FIM) of MHSA module with dumping parameter

$\qquad\mathbf{x}_{0}$: A guess of the update of tensors (may use the value
from previous batch)

\begin{algorithmic}[1]

\Function{FV}{$v$}
\State $\theta$=$g_{i}^{T}va_{i-1}$
\State $v_{1}$=$g_{i}\theta a_{i-1}^{T}+\gamma v$
\State \Return $v_{1}$ 

\EndFunction

\subparagraph*{Conjugate gradient (CG) iteration to approximate nature gradient}

\State $\boldsymbol{p}_{0}=\boldsymbol{r}_{0}=\boldsymbol{b}-F_{\gamma}\boldsymbol{x}_{0}$

\State $\rho_{0}=\left\Vert \boldsymbol{r}_{0}\right\Vert ^{2}$

\For{$k=0,1,2,\cdots,m$}

\State \quad{}$\boldsymbol{u}_{k}=FV(\boldsymbol{p}_{k})$

\State\quad{}$s_{k}=\boldsymbol{p}_{k}\cdot\boldsymbol{u}_{k}$

\State \quad{}$\alpha_{k}=\rho_{k}/s_{k}$

\State \quad{}$\boldsymbol{x}_{k+1}=\boldsymbol{x}_{k}+\alpha\boldsymbol{p}_{k}$

\State \quad{}$\boldsymbol{r}_{k+1}=\boldsymbol{r}_{k}-\alpha\boldsymbol{u}_{k}$

\State \quad{}$\rho_{k+1}=\left\Vert \boldsymbol{r}_{k+1}\right\Vert ^{2}$

\State\quad{}If $\rho_{k+1}$is sufficiently small, then exit

\State \quad{}$\beta_{k}=\rho_{k+1}/\rho_{k}$

\State \quad{}$\boldsymbol{p}_{k+1}=\boldsymbol{r}_{k+1}+\beta_{k}\boldsymbol{p}_{k}$

\EndFor

\State \Return

$\boldsymbol{x}_{k+1}$: The approximation of update tensor $d\theta$

\end{algorithmic}
\end{algorithm}

\subsection{Some discussions on more complex case} 
\subsubsection{Multi head and single head}
In the analysis above, the model we studied contains only one head. In practical applications, multiple headers are usually used. The complex structure of multi-head is just to reduce the total computational cost and get more parallelism. Some people believes that MHSA would combine the information from different representation sub-spaces. But as pointed in \cite{michel2019sixteen}: "we make the surprising observation that even if models have been trained using multiple heads, in practice, a large percentage of attention heads can be removed at test time without significantly
impacting performance. In fact, some layers can even be reduced to a single head.” So one head model double also be used in practical applications. And we only analyze one head model to simplify the derivation.

\subsubsection{Projection Matrix}
Many transformer models defines three learnable weight matrices: $\mathbf{W}^{Q}\in \mathbb{R}^{d\times d_{k}}$, $\mathbf{W}^{K}\in \mathbb{R}^{d\times d_{k}}$ and 
$\mathbf{W}^{V}\in \mathbb{R}^{d\times d_{v}}$. Then project input $ \mathbf{X}\in \mathbb{R}^{n\times d} $ to get query, key and value matrix:
\begin{equation}
\begin{split} \label{F:QKV}
     \mathbf{Q}&=\mathbf{X}\mathbf{W}^{Q}   \\
     \mathbf{K}&=\mathbf{X}\mathbf{W}^{K}   \\
     \mathbf{V}&=\mathbf{X}\mathbf{W}^{V}
\end{split}
\end{equation}
Our derivation is just the simplest case, all $\mathbf{W}^{Q},\mathbf{W}^{K},\mathbf{W}^{V}$ are identity matrix.
It should be pointed out that learnable $\mathbf{W}^{Q},\mathbf{W}^{K},\mathbf{W}^{V}$ would no longer corresponds to self attention. That is $\textbf{Q} \neq \textbf{K} \neq \textbf{V}$, since the weight in $\mathbf{Q},\mathbf{K},\mathbf{V}$ would change in the training process. For the analysis in section \ref{sec:brown}, we would introduce a more complex kernel function in a later paper.

\section{Conclusion and Prospect}
In this paper, we reveal the Brownian motion in the MHSA module of transformer model. This novel discovery would help to improve the model from a deeper perspective. For example, we present a new second-order optimizer(an iterative K-FAC algorithm). Our research reveals the connections between several distinct areas, such as machine learning, random walking and kernel functions. For a long time, the connection between these areas has been ignored. Now it's time to study the deep connection between them. We are testing some more ideas and more novel structure along this direction. We are doing a series of tests and would release the results soon. Some results are available at https://github.com/closest-git/DeepFormer.

\bibliographystyle{unsrt}  
\bibliography{references}  %%% Remove comment to use the external .bib file (using bibtex).


% George Kour and Raid Saabne.
% \newblock Real-time segmentation of on-line handwritten arabic script.
% \newblock In {\em Frontiers in Handwriting Recognition (ICFHR), 2014 14th
%   International Conference on}, pages 417--422. IEEE, 2014.

% George Kour and Raid Saabne.
% \newblock Fast classification of handwritten on-line arabic characters.
% \newblock In {\em Soft Computing and Pattern Recognition (SoCPaR), 2014 6th
%   International Conference of}, pages 312--318. IEEE, 2014.

% \bibitem{hadash2018estimate}
% Guy Hadash, Einat Kermany, Boaz Carmeli, Ofer Lavi, George Kour, and Alon
% \newblock Estimate and replace: A novel approach to integrating deep neural
%   networks with existing applications.
% \newblock {\em arXiv preprint arXiv:1804.09028}, 2018.

% \end{thebibliography}

\end{document}